%% file: acl_latex.tex
\pdfoutput=1

\documentclass[11pt]{article}

\usepackage{acl}
\usepackage{hyperref}
\usepackage{times}
\usepackage{latexsym}
\usepackage{multirow}
\usepackage{booktabs}
\usepackage{microtype}
\usepackage{multirow}
\usepackage{latexsym}
\usepackage{booktabs}
\usepackage{courier}
\usepackage{amsmath}
\usepackage{booktabs} 
\usepackage{siunitx} 
\usepackage{subfig}
\usepackage{placeins}
\usepackage{tikz}
\usepackage{hyperref}
\usepackage{svg}
\usepackage{pdfpages}

\usepackage[T1]{fontenc}

\usepackage[utf8]{inputenc}

\usepackage{microtype}

\usepackage{inconsolata}
\usepackage{soul}

\title{IITK at SemEval-2024 Task 4: Hierarchical Embeddings for Detection of Persuasion Techniques in Memes}

\author{Shreenaga Chikoti \qquad  Shrey Mehta \qquad Ashutosh Modi \\
Indian Institute of Technology Kanpur (IIT Kanpur)\\
\texttt{chikoti20@iitk.ac.in} \\ \texttt{ashutoshm@cse.iitk.ac.in} 
}

\begin{document}
\maketitle

\input{sections/abstract}
\input{sections/introduction}

\input{sections/background}

\input{sections/system}

\input{sections/experiments}
\input{sections/results}

\input{sections/conclusion}


\bibliography{custom}

\end{document}

%% file: sections/abstract.tex
\begin{abstract}
            Memes are one of the most popular types of content used in an online disinformation campaign. They are primarily effective on social media platforms since they can easily reach many users. Memes in a disinformation campaign achieve their goal of influencing the users through several rhetorical and psychological techniques, such as causal oversimplification, name-calling, and smear. The SemEval 2024 Task 4 \textit{Multilingual Detection of Persuasion Technique in Memes} on identifying such techniques in the memes is divided across three sub-tasks: ($\mathbf{1}$) Hierarchical multi-label classification using only textual content of the meme, ($\mathbf{2}$) Hierarchical multi-label classification using both, textual and visual content of the meme and ($\mathbf{3}$) Binary classification of whether the meme contains a persuasion technique or not using it's textual and visual content. This paper proposes an ensemble of Class Definition Prediction (CDP) and hyperbolic embeddings-based approaches for this task. We enhance meme classification accuracy and comprehensiveness by integrating HypEmo's hierarchical label embeddings \cite{chen2023labelaware}  and a multi-task learning framework for emotion prediction. We achieve a hierarchical F1-score of 0.60, 0.67, and 0.48 on the respective sub-tasks.
\end{abstract}

%% file: sections/introduction.tex
\section{Introduction} \label{sec:intro}



\begin{figure}[t]
\centering
    \includegraphics[scale = 0.1]{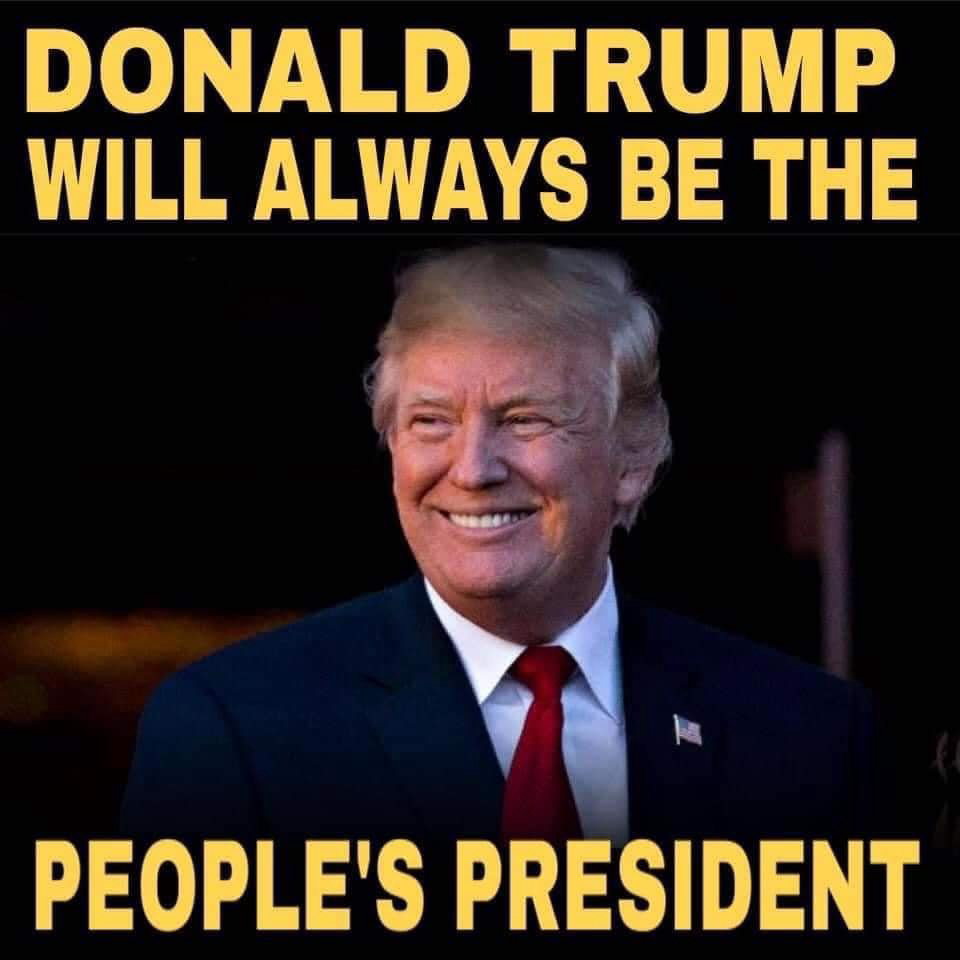} 
    \includegraphics[scale = 0.1]{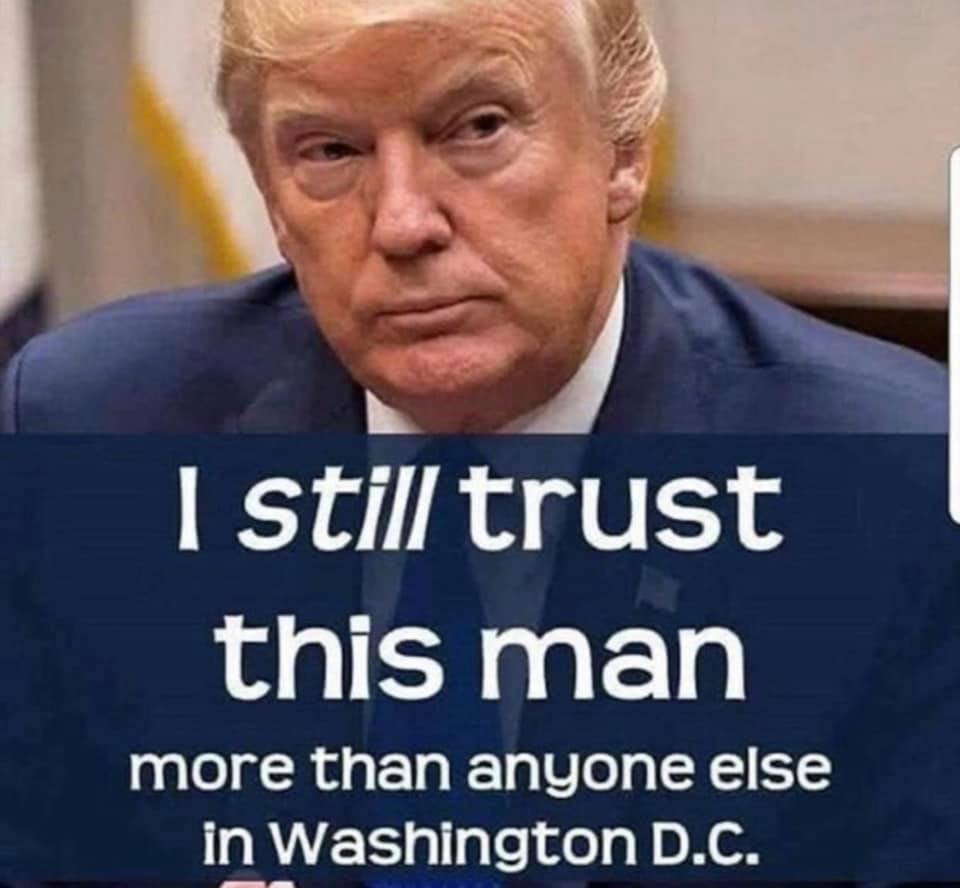} 
    \includegraphics[scale = 0.1]{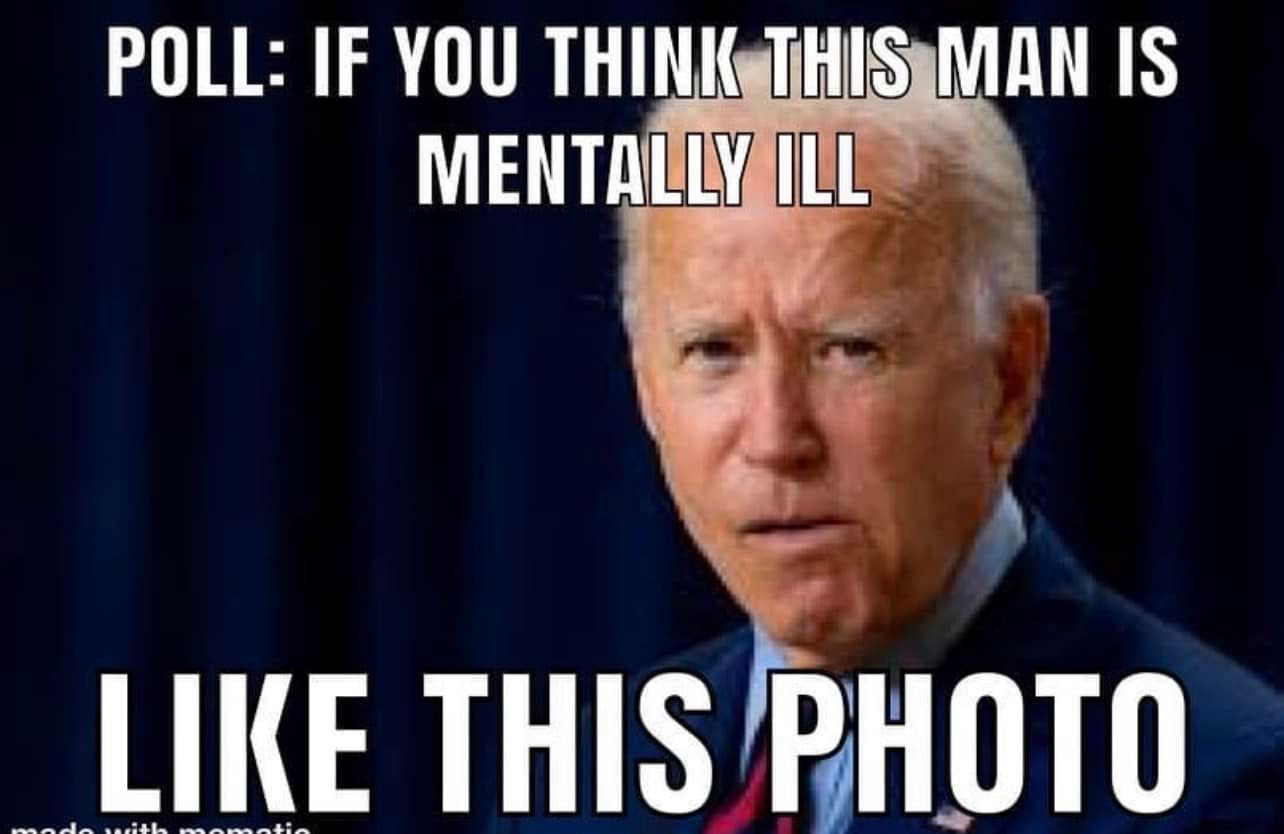} 
    \captionof{figure}{Sample set of memes showing the multi-modal setting}
     \label{fig:examples}
\end{figure}

Memes are popular among people of all age groups today through different social media platforms \cite{keswani-etal-2020-iitk-semeval,singh-etal-2020-newssweeper}. These memes help people know about the trends around them and can influence their decisions. Memes are one of the popular modes for spreading disinformation among people (examples in Figure \ref{fig:examples}), as studies have suggested that people tend to believe what they see frequently in such memes spread over the internet \cite{article}. As evidenced by research \cite{shu2017fake} during the 2016 US Presidential campaign, nefarious actors, including bots, cyborgs, and trolls, leveraged memes to evoke emotional reactions and propagate misleading narratives \cite{guo2020future}.

\begin{figure*}[t]
    \centering
    \includegraphics[scale=0.35]{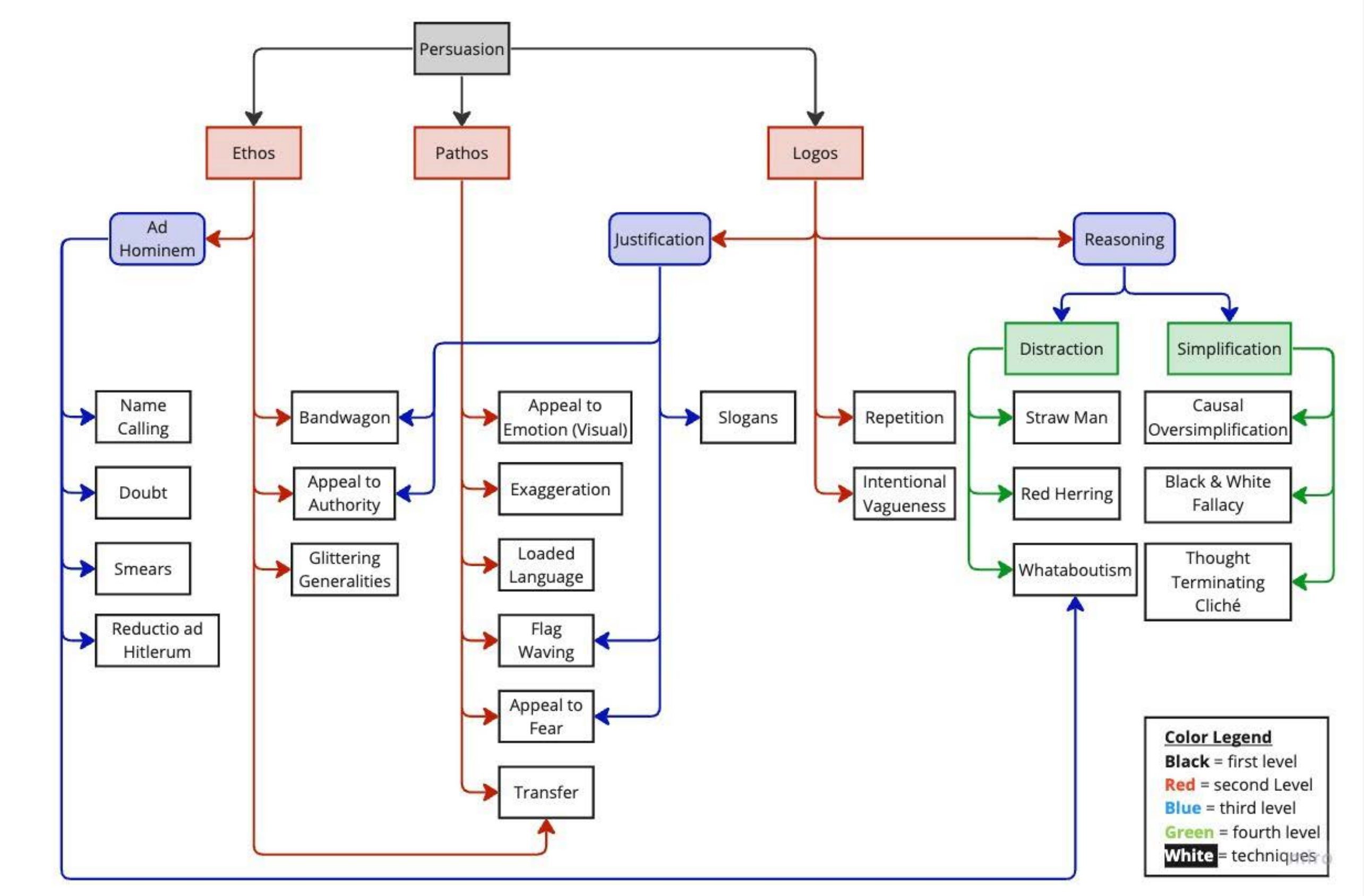}
    \caption{Taxonomy of persuasion techniques for sub-task $\mathbf{2}$ }
    \label{fig:taxonomy}
\end{figure*}

In this respect, SemEval-2024 Task 4 \cite{semeval2024task4} focuses on predicting the persuasive technique (from the visual and textual content) used in a meme across four different languages: English, Arabic, North Macedonian and Bulgarian. The task is divided into three sub-tasks: ($\mathbf{1}$) Hierarchical multi-label classification using only textual content of the meme, ($\mathbf{2}$) Hierarchical multi-label classification using both textual and visual content of the meme and ($\mathbf{3}$) Binary classification of whether the meme contains a persuasion technique or not using it's textual and visual content. The training data is provided for each sub-task but only in English. Taxonomy of various persuasion techniques (Figure \ref{fig:taxonomy}) and their respective definitions are provided. 

To address sub-task $\mathbf{1}$, we employed a dual approach involving definition-based modeling for each class and hierarchical classification using hyperbolic embeddings, as proposed in \citet{chen2023labelaware}. Based on hyperbolic embeddings, the method facilitates a nuanced classification of persuasion techniques by leveraging hierarchical structures. The incorporation of definition-based modeling allows for a dataset-agnostic approach, enhancing the precision of classification without reliance on hierarchical structures.

For sub-task $\mathbf{2}$, we augmented our methodology by integrating CLIP embeddings  \cite{radford2021learning} to capture essential features from memes' textual and visual components. This fusion of textual and visual information enables a more comprehensive analysis of meme content.

In addressing sub-task $\mathbf{3}$, we adopted an ensemble approach, leveraging transfer learning from both the DistilBERT \cite{sanh2019distilbert} and CLIP embeddings \cite{radford2021learning}. This ensemble technique enhances the robustness and effectiveness of our classification system by amalgamating insights from both pre-trained models. We release the code via GitHub.\footnote{\url{https://github.com/Exploration-Lab/IITK-SemEval-2024-Task-4-Pursuasion-Techniques}}

%% file: sections/background.tex
\section{Background} \label{sec:background}

The goal of propaganda is to enhance people's mindsets \cite{singh-etal-2020-newssweeper}, especially at the time of elections, where the trends in the media influence the votes of the people \cite{shu2017fake}. Propaganda uses psychological and rhetorical techniques to serve its purpose. Such methods include using logical fallacies and appealing to the audience's emotions. Logical fallacies are usually hard to spot since the argumentation, at first sight, might seem correct and objective. However, a careful analysis shows that the conclusion cannot be drawn from the premise without misusing logical rules \cite{gupta-sharma-2021-nlpiitr}. Another set of techniques uses emotional language to induce the audience to agree with the speaker only based on the emotional bond that is being created, provoking the suspension of any rational analysis of the argumentation \cite{article1}. 

Corpora development has been instrumental in advancing deception detection methodologies. \citet{rashkin2017truth} introduced the TSHP-17 corpus, providing document-level annotation across four classes: trusted, satire, hoax, and propaganda. However, their study on the classification task revealed limitations in the generalizability of n-gram-based approaches. Building on this, \citet{barron2019proppy} contributed the QProp corpus, which specifically targeted propaganda detection, employing a binary classification scheme of propaganda versus non-propaganda. Similarly, \citet{habernal2018adapting} developed a corpus annotated with fallacies, including \textit{ad hominem} and \textit{red herring}, directly relevant to propaganda techniques.

BERT-based variants have emerged as promising methodologies for classification tasks in tandem with corpus development. \citet{yoosuf2019fine} proposed a fine-tuning approach post-world-level classification using BERT, while \citet{fadel2019pretrained} presented a pre-trained ensemble model integrating BiLSTM, BERT, and RNN components. Further extending the capabilities of BERT, \citet{costa2023clac} advocated for a multilingual setup, employing translation to English before utilizing RoBERTa. Additionally, \citet{teimas2023detecting} proposed a hybrid technique combining CNN with DistilBERT for improved detection accuracy.

Exploring multimodal content, \citet{glenski2019multilingual} delved into multilingual multimodal deception detection, mainly focusing on hateful memes. Leveraging visual and textual content, they utilized fine-tuning techniques with state-of-the-art models like ViLBERT and VisualBERT and transfer learning-based approaches \cite{gupta2021volta}.

\section{Data Description}

The competition consisted of two different phases mainly the development phase which we refer to as the development set and for the development phase we were provided the training and validation sets for benchmarking our models

\begin{figure}[t]
    \centering
    \includegraphics[scale=0.35]{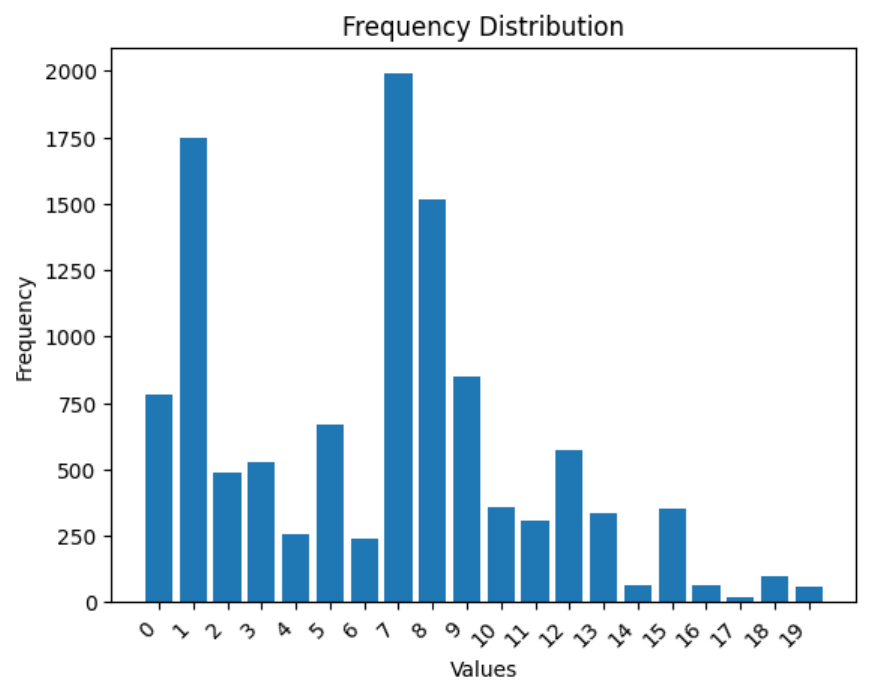}
    \caption{The frequency Distribution of Labels in the training dataset}
    \label{fig:dist1}
\end{figure}

\begin{figure}[t]
    \centering
    \includegraphics[scale=0.35]{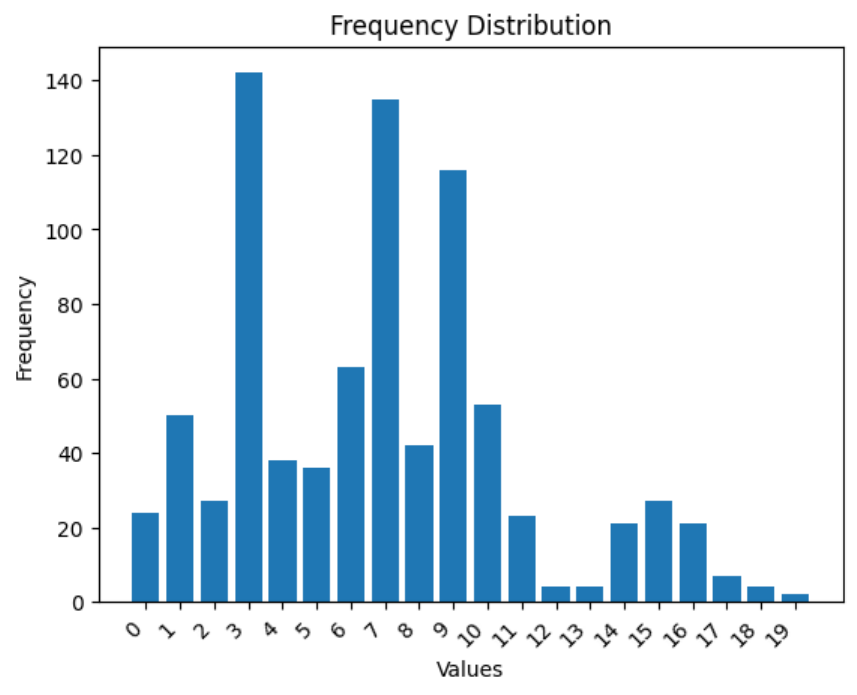}
    \caption{The frequency Distribution of Labels in the validation dataset}
    \label{fig:dist2}
\end{figure}

All three sub-tasks have different sets of memes split across training, validation and Develepmont sets as shown in Table \ref{table:stats}. 
We have also plotted the Distribution of the labels across the Figure \ref{fig:dist1} training data and the Figure \ref{fig:dist2} validation data.

Our analysis used a dictionary to map various rhetorical techniques to numerical values for plotting. This dictionary is as follows:

\begin{table}[t]
\centering
\tiny
\begin{tabular}{@{}lc@{}}
\toprule
\textbf{Persuasion Technique} & \textbf{Number mapped to} \\
\midrule
Presenting Irrelevant Data (Red Herring) & 0\\
Bandwagon & 1 \\
Smears & 2\\
Glittering generalities (Virtue) & 3\\
Causal Oversimplification & 4\\
Whataboutism & 5\\
Loaded Language & 6  \\
Exaggeration/Minimisation & 7\\
Repetition & 8\\
Thought-terminating cliché & 9   \\
Name calling/Labeling & 10  \\
Appeal to authority & 11\\
Black-and-white Fallacy/Dictatorship & 12\\
\begin{tabular}{@{}l@{}}Obfuscation, Intentional vagueness, \\ Confusion (Straw Man)\end{tabular} & 13\\
Reductio ad hitlerum & 14\\
Appeal to fear/prejudice &  15\\
\begin{tabular}{@{}l@{}}Misrepresentation of Someone's \\ Position (Straw Man)\end{tabular} & 16 \\
Flag-waving & 17 \\
Slogans & 18 \\
Doubt & 19 \\
\bottomrule
\end{tabular}%
\medskip
\caption{Dictionary Mapping for different persuasion techniques for Subtask 1}
\label{table:metrics1}
\end{table}

\begin{figure}
    \centering
    \includegraphics[scale=0.20]{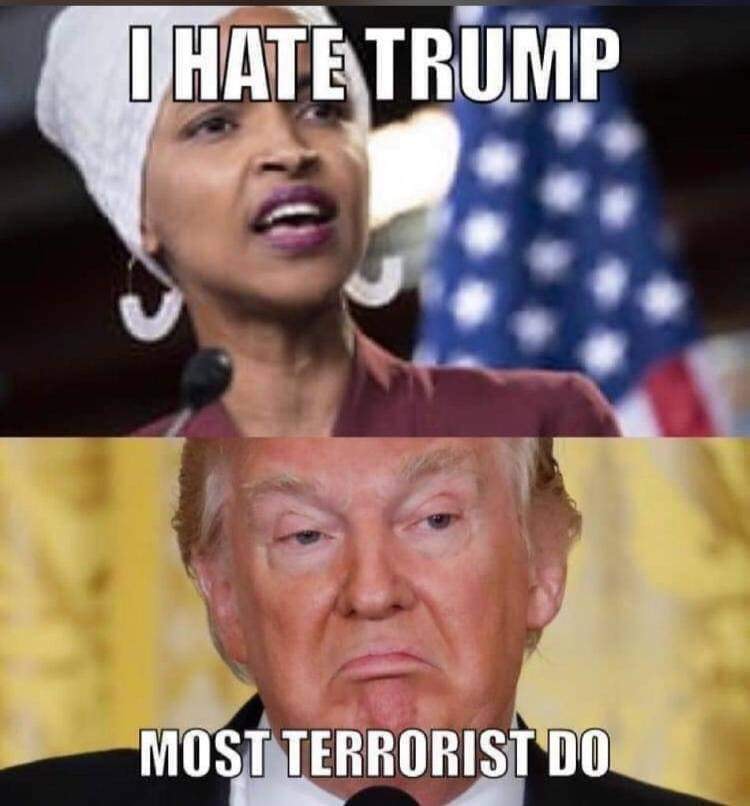}
    \caption{The meme sarcastically suggests that individuals who oppose Trump are being unfairly equated with terrorists, highlighting the absurdity of such comparisons. Two persuasion
techniques are used: (i) \textit{Loaded Language},
and (ii) \textit{Name calling}  that can be inferred from the text and the visual content.}

\end{figure}

\begin{table}[t]
\centering
\resizebox{\columnwidth}{!}{
\begin{tabular}{@{}lccc@{}}
\toprule
\textbf{Sub-task} & \textbf{Train Data} & \textbf{Validation Data} & \textbf{Development Data} \\
\midrule
 Sub-task 1 & 7000 & 500 & 1000 \\
 Sub-task 2 & 7000 & 500 & 1000 \\
 Sub-task 3 & 1200 & 300 & 500 \\
\bottomrule
\end{tabular}%
}
\caption{Distribution of data across sub-tasks }
\label{table:stats}
\end{table}

%% file: sections/system.tex
\section{System overview} \label{sec:system}

The proposed system for all the sub-tasks involves task-specific modifications made to the BERT model and earlier proposed works including CLIP Model \cite{radford2021learning}, Class Definition based Emotion Predictions \cite{singh2021finegrained,singh2023TFACC} and HypEmo model \cite{chen2023labelaware} (described below).

\subsection{Data Pre-processing}
To ensure consistency and standardization, we begin by pre-processing the text. This involves removing newline characters, commas, numerical values, and other special characters. Additionally, the entire text is converted to lowercase. In our approach, we leverage the Development (Dev) and Training sets, focusing solely on samples containing non-zero classes.

\subsection{Sub-task 1: Hierarchical Multi-label Text Classification}

We present a novel approach to meme classification, drawing upon the methodologies of two key frameworks: HypEmo and a multi-task learning model focused on emotion definition modeling.

HypEmo \cite{chen2023labelaware} utilizes pre-trained label hyperbolic embeddings to capture hierarchical structures effectively, particularly in tree-like formations. Initially, the hidden state of the [CLS] token from the RoBERTa backbone model is projected using a Multi-Layer Perceptron (MLP). Subsequently, an exponential map is applied to project it into hyperbolic space. The distance from pre-trained label embeddings is the weight for the cross-entropy loss function, enhancing the model's sensitivity to label relationships.

To implement the HypEmo architecture, we transform the Directed Acyclic Graph (DAG) (Figure  \ref{fig:taxonomy}) into a tree structure. This involves duplicating children with multiple parents, resulting in distinct embeddings for each label. For example, a sentence with various labels is converted into separate samples, each assigned one label. Utilizing the Poincaré hyperbolic entailment cones model \cite{ganea2018hyperbolic} with 100 dimensions, the constructed tree undergoes training, with predictions generated via softmax. Peaks are identified through Z-score analysis associated with each class, with thresholds set accordingly. 

\citet{singh2021finegrained,singh2023TFACC} have introduced a complementary approach focusing on emotion prediction through a multi-task learning framework. This model incorporates auxiliary tasks, including masked language modeling (MLM) and class definition prediction, to enhance the understanding of emotional concepts. In our setup, class definitions are merged using a [SEP] token, with the model trained to predict whether the conjoined definition matches the actual definition. Binary cross-entropy loss is employed for this task, along with MLM for fine-tuning the model. Additionally, binary cross-entropy loss is used for each class during training. We utilize class definitions provided by the meme classification competition for the auxiliary task of class-definition prediction.

Finally, we merge the predictions generated by both models
(HypEmo, Fine-grained class-definition based model) to compute the final predictions. This integrated approach aims to leverage the strengths of each framework, enhancing the accuracy and comprehensiveness of meme classification outcomes.

\subsection{Sub-task 2: Hierarchical Multi-label Text and Image Classification}

\begin{figure}[t]
    \centering
     \includegraphics[scale = 0.4]{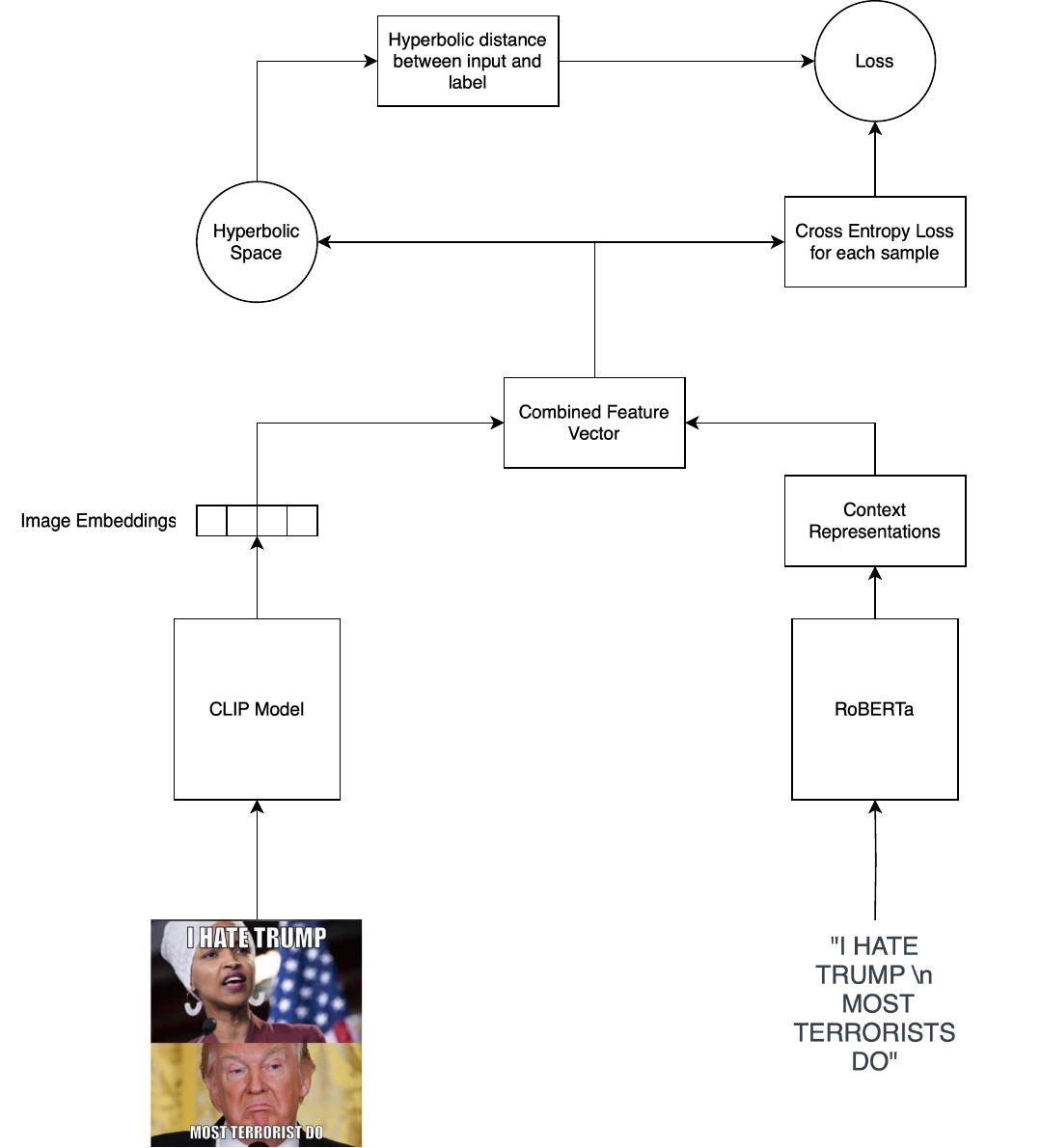}
    \caption{Proposed architecture for sub-task $\mathbf{2}$}
    \label{fig:enter-label}
\end{figure}

We model this sub-task by experimenting with using an ensemble of HypEmo \cite{chen2023labelaware} and the class definition-based multi-task learning model \cite{singh2021finegrained,singh2023TFACC} for the textual content of the meme and using the CLIP model \cite{radford2021learning} embeddings for extracting the relevant features from the visual content of the meme. We construct a similar DAG structure for sub-task 1 and generate the hyperbolic embeddings. The image embeddings obtained from the CLIP model are concatenated with the embeddings generated for the textual contents before sending the combined feature vector for training. Then, the model is trained, and the predictions are generated using the softmax activation function. The Z-score analysis is done on the resulting predictions to make the classification, similar to task 1. An overview of the architecture of the modified HypEmo model is shown in Figure 7.

\subsection{Sub-task 3: Binary Text and Image Classification}
In this task, we must classify whether a meme contains a persuasion technique based on its textual and visual content. We use the pre-trained BERT\textsubscript{BASE} model \cite{devlin2019bert} and the Convolution Neural Network (CNN) \cite{oshea2015introduction} layer to extract the features from the text and image, respectively. We attach a feed-forward $[CLS]$ token embedding along with two linear layers connected by the $sigmoid$ activation function in between, which generates the sentence embeddings corresponding to the textual content in the meme. We use a network of four CNN layers connected through the ReLU activation function, which progressively extracts features from the input image. Max pooling layers are used to down-sample the feature maps, increasing robustness to minor variations. The resultant image embeddings are concatenated with the sentence embeddings, and a linear classifier is applied to the combined feature vector with the $sigmoid$ activation function. We use the binary cross-entropy loss function to train the model and tune the hyperparameters on the validation set. An overview of the model architecture is shown with an example in Figure 8.

\begin{figure}[t]
    \centering
     \includegraphics[scale = 0.37]{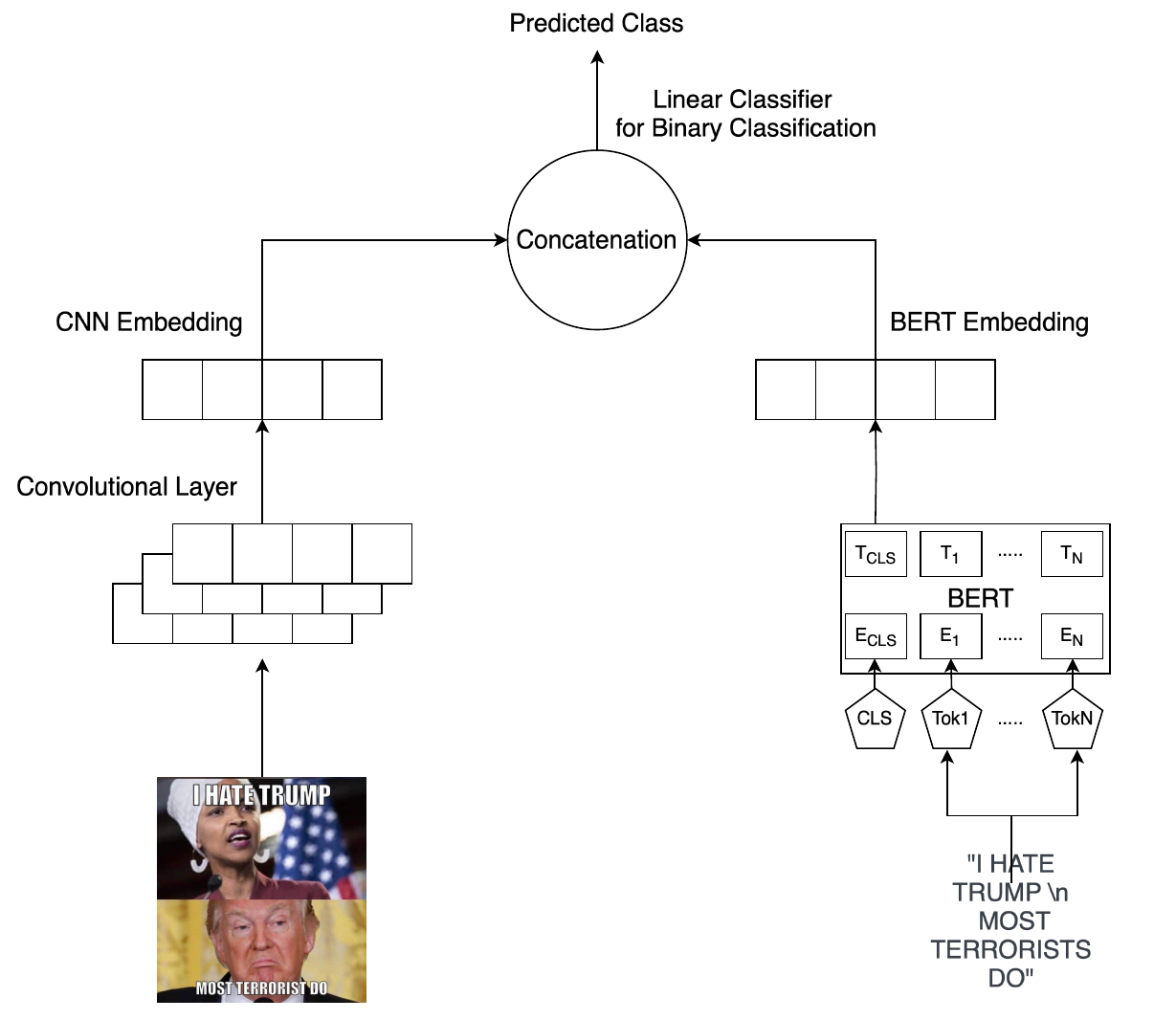}
    \caption{Proposed architecture for sub-task $\mathbf{3}$}
    \label{fig:enter-label}
\end{figure}

Since the training data is in a 2:1 ratio for the ``persuasive'' (positive, labeled as 1) and ``not-persuasive'' (negative, labeled as 0) class, which leads to an imbalance in the dataset, we use the weighted binary cross entropy loss function as shown below:
\begin{equation*}
\begin{split}
    L(\textbf{x}, \textbf{y}) &= -\frac{1}{N} \sum_{1}^{N}(w*y_i*log(x_i) \\ &+ (1-w)*(1-y_i)*log(1-x_i)) 
\end{split}
\end{equation*}
\begin{equation*}
        w = \frac{1}{f}(K-f)
\end{equation*}

where $N$ is the batch size, $i$ is the index of the $i^{th}$ batch element, $f$ is the frequency of the positive class, $\textbf{x}$ is the output of the last $sigmoid$ layer, $\textbf{y}$ is the vector of the ground truth labels, and $K$ is the total size of the training dataset. Finally, by choosing the one with a higher probability, we use the output probabilities of the final $sigmoid$ layer to predict whether a persuasion technique is present in the meme.

%% file: sections/experiments.tex
\section{Experimental setup} \label{sec:experiments}






\subsection{Implementation Details}

We have used the official PyTorch implementation \cite{paszke2019pytorch} for implementing all the models across sub-tasks. We have used the HypEmo\footnote{HypEmo, https://github.com/dinobby/hypemo} model and the Class Definition Prediction (CDP)\footnote{CDP, https://github.com/Exploration-Lab/FineGrained-Emotion-Prediciton-Using-Definitions} model for generating the hyperbolic embeddings and class-definition based features of the textual contents, respectively and the CLIP\footnote{CLIP, https://github.com/openai/CLIP} mainly the 'clip-ViT-B-32' model for generating embeddings for the visual features of the meme. Some portions of the test set have languages other than English for testing purposes. Since the models described earlier were trained in English, we translated the non-English data into English language using the implementation of the OPUS-MT model \cite{tiedemann-thottingal-2020-opus} from the HuggingFace\footnote{OPUS-MT, https://huggingface.co/Helsinki-NLP/opus-mt-bg-en} library and inference was done on the translated text. We created an ensemble of classes predicted by all the models and took a union of the predicted labels to produce the final predicted set of labels to which the meme belonged.

We have used the data in the same ratio provided in the task to train the models. We combine the train validation dataset for training in each subtask and test it in the four languages.

\subsection{Evaluation Metrics}
Sub-tasks 1 and 2 depend on a hierarchy, as shown in Figure \ref{fig:taxonomy}. Hierarchical-F1 \cite{inproceedings} is used as the evaluation metric for these two sub-tasks. In these two, the gold label is always a leaf node of the DAG, considering the hierarchy in Figure \ref{fig:taxonomy} as a reference. However, any node of the DAG can be a predicted label with:
\begin{itemize}
    \item If the prediction is a leaf node and it is the correct label, then a full reward is given. For example, \textit{ Red Herring} is predicted and is the gold label.
    \item If the prediction is NOT a leaf node and an ancestor of the correct gold label, then a partial reward is given (the reward depends on the distance between the two nodes). For example, if the gold label is Red Herring and the predicted label is \textit{Distraction} or \textit{Appeal to Logic}.
    \item If the prediction is not an ancestor node of the correct label, then a null reward is given. For example, if the gold label is\textit{ Red Herring} and the predicted label is\textit{ Black and White Fallacy} or\textit{ Appeal to Emotions.}
\end{itemize}
Sub-task 3 uses macro-F1 as the evaluation metric for the binary classification task. This ensures equal importance to the "persuasion technique present" and "no persuasion technique" classes, regardless of potential data imbalance.

%% file: sections/results.tex
\section{Results} \label{sec:results}

\begin{table}[t]
\centering
\tiny
\begin{tabular}{@{}lccc@{}}
\toprule
\textbf{Technique} & \textbf{Arabic} & \textbf{Bulgarian} & \textbf{North Macedonian} \\
\midrule
Presenting Irrelevant Data (Red Herring) &0.    & 0.    & 0.    \\
Bandwagon & 0.    & 0.    & 0. \\
Smears &0.67  & 0.84  & 0.90  \\
Glittering generalities (Virtue) & 0.29 & 0.10  & 0. \\
Causal Oversimplification &0.    & 0.    & 0.    \\
Whataboutism &0.    & 0.05  & 0.    \\
Loaded Language &0.41  & 0.62  & 0.37  \\
Exaggeration/Minimisation &0.  & 0.    & 0.    \\
Repetition &0.50  & 0.34  & 0.    \\
Thought-terminating cliché &0.  & 0.19  & 0.    \\
Name calling/Labeling &0.44  & 0.45  & 0.49  \\
Appeal to authority &0. & 0.30  & 0.31  \\
Black-and-white Fallacy/Dictatorship &0. & 0.06  & 0. \\
\begin{tabular}{@{}l@{}}Obfuscation, Intentional vagueness, \\ Confusion (Straw Man)\end{tabular} &0.    & 0. & 0. \\
Reductio ad hitlerum &0. & 0. & 0. \\
Appeal to fear/prejudice &0.04 & 0.21  & 0.1  \\
\begin{tabular}{@{}l@{}}Misrepresentation of Someone's \\ Position (Straw Man)\end{tabular} &0.  & 0.  & 0.  \\
Flag-waving &0.  & 0.33  & 0.  \\
Slogans &0. & 0.43  & 0.16  \\
Doubt &0.  & 0.15  & 0.11  \\
Transfer & 0. & 0.48 & 0.61 \\
Appeal to (Strong) Emotions & 0. & 0.18 & 0.09 \\
\bottomrule
\end{tabular}%
\medskip
\caption{Macro F1 scores for different persuasion classes for the given languages for Subtask 2}
\label{table:metrics}
\end{table}

\begin{table}[t]
\centering
\tiny
\begin{tabular}{@{}lccc@{}}
\toprule
\textbf{Technique} & \textbf{Arabic} & \textbf{Bulgarian} & \textbf{North Macedonian} \\
\midrule
Presenting Irrelevant Data (Red Herring) &0.    & 0.    & 0.    \\
Bandwagon & 0.    & 0.    & 0. \\
Smears &0.33  & 0.18  & 0.17  \\
Glittering generalities (Virtue) & 0. & 0.07  & 0. \\
Causal Oversimplification &0.    & 0.    & 0.    \\
Whataboutism &0.    & 0.08  & 0.    \\
Loaded Language &0.39  & 0.62  & 0.55  \\
Exaggeration/Minimisation &0.11  & 0.    & 0.    \\
Repetition &0.40  & 0.36  & 0.    \\
Thought-terminating cliché &0.  & 0.28  & 0.    \\
Name calling/Labeling &0.39  & 0.58  & 0.54  \\
Appeal to authority &0. & 0.38  & 0.22  \\
Black-and-white Fallacy/Dictatorship &0. & 0.04  & 0. \\
\begin{tabular}{@{}l@{}}Obfuscation, Intentional vagueness, \\ Confusion (Straw Man)\end{tabular} &0.    & 0. & 0. \\
Reductio ad hitlerum &0. & 0. & 0. \\
Appeal to fear/prejudice &0. & 0.05  & 0.  \\
\begin{tabular}{@{}l@{}}Misrepresentation of Someone's \\ Position (Straw Man)\end{tabular} &0.  & 0.  & 0.  \\
Flag-waving &0.  & 0.29  & 0.  \\
Slogans &0. & 0.37  & 0.04  \\
Doubt &0.25  & 0.16  & 0.1  \\
\bottomrule
\end{tabular}%
\medskip
\caption{Macro F1 scores for different persuasion classes for the given languages for Subtask 1}
\label{table:metrics1}
\end{table}

We conducted several experiments across all the sub-tasks, and the detailed information can be seen in Table \ref{table:metrics},Table \ref{table:metrics1},Table \ref{table:metrics2},Table \ref{table:metrics3} and Table \ref{table:metrics_contrast4}.

\begin{table}[t]
\centering
\scriptsize
\begin{tabular}{@{}lcccc@{}}
\toprule
\textbf{Language} & \textbf{\begin{tabular}{@{}c@{}}Base \\ F1\end{tabular}} & \textbf{\begin{tabular}{@{}c@{}}Hierarchical \\ F1\end{tabular}} & \textbf{\begin{tabular}{@{}c@{}}Hierarchical \\ Precision\end{tabular}} & \textbf{\begin{tabular}{@{}c@{}}Hierarchical \\ Recall\end{tabular}} \\
\midrule
English&0.37 &  0.60 & 0.53 & 0.69 \\
Arabic& 0.37 & 0.42 & 0.32 & 0.60 \\
Bulgarian& 0.28 & 0.48 & 0.40 & 0.62 \\
 \begin{tabular}{@{}l@{}}North- \\ Macedonian\end{tabular}& 0.30 & 0.41 & 0.33 & 0.56 \\
\bottomrule
\end{tabular}%
\medskip
\caption{Hierarchical-F1 scores computed across four languages of the test set for sub-task 1. Base F1 score here is the Baseline F1 score }
\label{table:metrics2}
\end{table}

\begin{table}[t]
\centering
\scriptsize
\begin{tabular}{@{}lcccc@{}}
\toprule
\textbf{Language} & \textbf{\begin{tabular}{@{}c@{}}Baseline \\ F1\end{tabular}} & \textbf{\begin{tabular}{@{}c@{}}Hierarchical \\ F1\end{tabular}} & \textbf{\begin{tabular}{@{}c@{}}Hierarchical \\ Precision\end{tabular}} & \textbf{\begin{tabular}{@{}c@{}}Hierarchical \\ Recall\end{tabular}} \\
\midrule
English&0.44 &  0.67 & 0.67 & 0.67 \\
Arabic& 0.57 & 0.53 & 0.50 & 0.57 \\
Bulgarian& 0.50 & 0.65 & 0.66 & 0.63 \\
\begin{tabular}{@{}l@{}}North- \\ Macedonian\end{tabular}& 0.55 & 0.67 & 0.72 & 0.62 \\
\bottomrule
\end{tabular}%
\medskip
\caption{Hierarchical-F1 scores calculated for four languages within the test set for sub-task 2, with Base-F1 denoting the Baseline F1 score depicted on the leaderboard}
\label{table:metrics3}
\end{table}

\begin{table}[t]
\centering
\scriptsize
\begin{tabular}{@{}lcccc@{}}
\toprule
\textbf{Model} & \textbf{English} & \textbf{Arabic} & \textbf{Bulgarian} & \textbf{North-Macedonian}\\
\midrule
BERT  & 0.55 & 0.39	 & 0.40 & 0.36\\
RoBERTa  & 0.60 & 0.37 & 0.45 & 0.38 \\
HypEmo  & 0.55 & 0.43 & 0.42 & 0.39\\
CDP  & 0.59 & 0.40 & 0.48 & 0.43 \\
\begin{tabular}{@{}l@{}}HypEmo \\ + CDP \\ (Union)\end{tabular} & \textbf{0.60} &\textbf{ 0.42} &\textbf{ 0.48} & \textbf{0.41} \\
\bottomrule
\end{tabular}%
\medskip
\caption{Hierarchical-F1 scores calculated for four languages within the test set for sub-task 1 across different models}
\label{table:metrics3}
\end{table}

\begin{table}[t]
\centering
\scriptsize
\begin{tabular}{@{}lcccc@{}}
\toprule
\textbf{Model} & \textbf{English} & \textbf{Arabic} & \textbf{Bulgarian} & \textbf{North-Macedonian} \\
\midrule
\begin{tabular}{@{}l@{}}HypEmo \\ (Without CLIP) \end{tabular}  & 0.63 &  0.511 & 0.58  & 0.63 \\
\begin{tabular}{@{}l@{}}HypEmo \\ (With CLIP) \end{tabular}   & 0.63 & 0.49 & 0.59 & 0.62 \\
CDP & 0.64 & 0.51 & 0.62 & 0.65  \\
\begin{tabular}{@{}l@{}}HypEmo \\ + CDP \\ (Union)\end{tabular} & \textbf{0.67} & \textbf{0.53} &\textbf{0.65} &\textbf{ 0.67}  \\
\bottomrule
\end{tabular}%
\medskip
\caption{Hierarchical-F1 scores calculated for four languages within the test set for sub-task 2 across different models}
\label{table:metrics3}
\end{table}

\begin{table}[h!]
\centering
\scriptsize
\begin{tabular}{@{}lcc@{}}
\toprule
\textbf{Language} & \textbf{Base F1} & \textbf{Macro-F1} \\
\midrule
English & 0.25 & 0.49  \\
Arabic & 0.23 & 0.47 \\
North Macedonian & 0.09 & 0.49 \\
Bulgarian & 0.16 & 0.48 \\
\bottomrule
\end{tabular}%
\medskip
\caption{Macro-F1 scores computed across 4 languages of the test set for sub-task 3.}
\label{table:metrics_contrast4}
\end{table}

\begin{table}[h!]
\centering
\scriptsize
\begin{tabular}{@{}lc@{}}
\toprule
\textbf{Sub-task} & \textbf{Ranking}  \\
\midrule
English-Sub-task1 & 21 \\
English-Sub-task2 & 10\\
English-Sub-task3 & 19\\
Bulgarian-Sub-task1 & 14\\
Bulgarian-Sub-task2 & 8\\
Bulgarian-Sub-task3 & 11\\
North Macedonian-Sub-task1 & 13\\
North Macedonian-Sub-task2 & 7\\
North Macedonian-Sub-task3 & 7\\
Arabic-Sub-task1 & 4\\
Arabic-Sub-task2 & 6\\
Arabic-Sub-task3 & 13\\
\bottomrule
\end{tabular}%
\medskip
\caption{Leaderboard position of our team in the competition in each sub-task}
\label{table:metrics_contrast5}
\vspace{-4mm}
\end{table}

For Task 1, we started experimenting with the BERT and RoBERTa models, achieving a hierarchical F1 score of $0.55$ and $0.60$ on the test set of the English language. But, in this approach, we did not take the hierarchy and the definitions of the classes into consideration. We tried to accommodate that using the combination of HypEmo and CDP models.

For the HypEmo model, the model was trained to prioritize higher-level labels in the Directed Acyclic Graph (DAG). During this process, we explored two options: eliminating children when the model predicted the parent label and retaining the children. We observed a significant impact on the hierarchical F1 score, with the first formulation yielding $0.45$ F1 and the second approach resulting in  $0.59$ on the test set. We also tried to predict the labels utilizing only the definitions of the classes, using the CDP model, which yielded a hierarchical F1 score of $0.57$ and $0.59$ on the dev set and the test set, respectively.

For constructing an ensemble, one approach considered concatenating embeddings or softmax predictions from both models for further classification using a neural network. However, this approach was not viable due to limited samples for generalization. The most effective model emerged from utilizing the ensemble with fine-tuning of hyperparameters. Combining predictions from both models yielded a hierarchical F1 score of $0.60$.

Table \ref{table:metrics3} shows that the best generalizability across all tasks is achieved via the HypEmo + CDP(Union) for subtask1.

For sub-task 2, we trained the model from scratch after including the two labels used in the ensemble used in sub-task 1 and changed the feature embeddings being trained by considering the features from the visual content. However, as you can see, there is very little to no difference between the results using CLIP and not using CLIP. We can also see that, unlike the first subtask, they perform better due to more data. 

We can see the F1-score analysis tables for each subtask, i.e., in    Table \ref{table:metrics1}, Table \ref{table:metrics} for subtask1 and subtask2.

For sub-task 3, we trained the model on an ensemble of BERT and CNN models to consider the textual and visual features. It was seen that the ensemble performs just slightly better than using the BERT model, that is, considering only the textual cues. Visual cues are considered significantly when persuasion techniques like $Smears$ are used, as seen in sub-task 2. For the rest of the persuasion techniques, the visual cues were seen not to make a significant impact on the classification task. On the gold labels of the dev set, the ensemble gave a macro-F1 score of $0.67$, which is a slight improvement from the BERT model, which showed a macro-F1 score of $0.63$ on the dev set.

%% file: sections/conclusion.tex
\section{Conclusion} \label{sec:conclusion}

Detection of persuasion techniques in memes is seen in a multi-modal setting in this task, but the significant features are drawn from the textual cues in the memes, which can be seen in the results of sub-tasks 1 and 2. The CLIP and other visual language models still need considerable development, and visual cues are helpful for only specific input-output pairs. Identifying whether a persuasion technique is present in the meme but does not apply to the multi-label classification task can be beneficial. Also, we have used a basic ensemble of the latest works in this area and modified them for task-specific requirements. Still, other complex architectures can be explored to get better results.